\documentclass[]{article}

\PassOptionsToPackage{numbers}{natbib}
     \usepackage[preprint]{neurips_2019}

\usepackage[utf8]{inputenc} % allow utf-8 input
\usepackage[T1]{fontenc}    % use 8-bit T1 fonts
\usepackage{hyperref}       % hyperlinks
\usepackage{url}            % simple URL typesetting
\usepackage{booktabs}       % professional-quality tables
\usepackage{amsfonts}       % blackboard math symbols
\usepackage{nicefrac}       % compact symbols for 1/2, etc.
\usepackage{microtype}      % microtypography
\usepackage{xcolor}
\usepackage{graphicx}
\usepackage{amsmath}
\usepackage{times}
\usepackage{latexsym}
\usepackage{multirow}
\usepackage{amsmath}
\usepackage{url}
\usepackage{makecell}
\usepackage{array}
\usepackage[ruled,linesnumbered]{algorithm2e}
\usepackage{subfigure}
\usepackage{enumitem}

\title{Say What I Want: Towards the Dark Side \\of Neural Dialogue Models}

\author{%
  Haochen Liu \\
  Michigan State University\\
  \texttt{liuhaoc1@msu.edu} \\
  \And
  Tyler Derr \\
  Michigan State University \\
  \texttt{derrtyle@msu.edu} \\
  \AND
  Zitao Liu \\
  TAL AI Lab \\
  \texttt{liuzitao@100tal.com} \\
  \And
  Jiliang Tang \\
  Michigan State University\\
  \texttt{tangjili@msu.edu} \\
}
\begin{document}

\maketitle

\begin{abstract}
Neural dialogue models have been widely adopted in various chatbot applications because of their good performance in simulating and generalizing human conversations.
However, there exists a dark side of these models -- due to the vulnerability of neural networks, a neural dialogue model can be manipulated by users to say what they want, which brings in concerns about the security of practical chatbot services.
In this work, we investigate whether we can craft inputs that lead a well-trained black-box neural dialogue model to generate targeted outputs.
We formulate this as a reinforcement learning (RL) problem and train a \textbf{Reverse Dialogue Generator} which efficiently finds such inputs for targeted outputs.
Experiments conducted on a representative neural dialogue model show that our proposed model is able to discover %find out 
such desired inputs in a considerable portion of cases.
Overall, our work reveals this weakness of neural dialogue models and may prompt further researches of developing corresponding solutions to avoid it. % \blue{last sentence seemed a little awkward}
\end{abstract}

\section{Introduction}

% {\bf why you use these manual references???}
% \red{I noticed the template uses manual references.. But I will change them.}

Dialogue system, also known as conversational AI, which aims to conduct human-like conversations with users, is receiving increasing attention from both the industry and the academic research community.
In the past, such systems either rely on intricate hand-crafted rules \cite{DBLP:journals/cacm/Weizenbaum66, DBLP:conf/interspeech/GoddeauMPSB96}, or depend on a complicated processing pipeline including a series of functional modules \cite{DBLP:reference/crc/MottLB04}.
Meanwhile, retrieval-based methods \cite{DBLP:conf/acl/Banchs12, DBLP:conf/iva/AmeixaCFQ14, DBLP:conf/nips/LuL13, DBLP:journals/corr/HuLLC15}, which search a suitable response from a repository given the query, are also adopted in many application scenarios.
These methods are able to provide natural, human-like responses, but fail to generate novel responses out of the range of the repository \cite{DBLP:journals/corr/SongYLZZ16}.
Recently, researchers begin to involve deep learning techniques in building fully data-driven and end-to-end dialogue systems \cite{DBLP:journals/ftir/GaoGL19}, which are referred to as neural dialogue models. 

% According to the practical usage of dialogue systems, they can be categorized into two groups: task-oriented systems, which assist users to complete specific tasks such as booking a flight ticket, and chatbots, which are designed to chit-chat with human users in open domains for entertainment [1].
% In this paper, we focus on the latter case, chatbots.

% \blue{I'm not sure if the two types should be mentioned here... the transition from this paragraph to the second isn't so smooth. I'm thinking maybe after the first sentence it could say something about in the past they used, other methods, but now the focus has switched to deep learning due to their surprising and far superior performance. This can then lead into the second paragraph that starts to discuss the details of the neural model performance (i.e., seq2seq, training, etc). Then after the second paragraph, the focus switches to the main topic of the dark side and problems/vulnerability. }

Based on the Seq2Seq framework \cite{DBLP:conf/nips/SutskeverVL14}, these neural models \cite{DBLP:conf/naacl/SordoniGABJMNGD15, DBLP:journals/corr/VinyalsL15, DBLP:conf/acl/ShangLL15, DBLP:conf/aaai/SerbanSBCP16, DBLP:conf/aaai/SerbanSLCPCB17} have achieved surprising performances and gradually dominate the field of dialogue generation.
First, these models are easy to train.
Instead of designing complicated rules or modularized pipelines, the models can learn the mapping between queries and responses automatically from massive existing dialogue pairs \cite{DBLP:journals/corr/VinyalsL15}.
Second, given that the neural models are trained on large-scale human conversation data, they show a strong generalization ability that they can handle open-domain conversations rather than restricting the topics in a narrow domain \cite{DBLP:journals/corr/VinyalsL15, DBLP:journals/corr/abs-1812-08989}.
In addition, neural dialogue models can provide fluent and smooth responses, and show the intelligence of performing simple common sense reasoning \cite{DBLP:journals/corr/VinyalsL15}.
Since neural dialogue models achieve a breakthrough of conducting reasonable and engaging human-like conversations, they are widely adopted by the industry as a core component of practical chatbot applications, such as Microsoft XiaoIce \cite{DBLP:journals/corr/abs-1812-08989}.
% Thus, neural dialogue models achieve a breakthrough of conducting reasonable and engaging human-like conversations, and they are widely adopted by the industry to build practical chatbot applications, such as Microsoft XiaoIce \cite{DBLP:journals/corr/abs-1812-08989}. \blue{last sentence is a little awkward}

% \blue{In general this paragraph feels too long. (I'll have to look at other neurips papers to see how long the introductions usually are though as I'm not too familiar with this single column format.) I think it might not be necessary to include the history of dialogue models here. }

While the research community is delighted with the success of neural dialogue models, there is a dark side of these models. 
Given that the internal mechanisms of neural networks are not explicitly interpretable, neural dialogue models are vulnerable.  For example, they may have unpredictable behaviors with regard to some well-crafted inputs \cite{DBLP:journals/corr/SzegedyZSBEGF13, DBLP:journals/corr/abs-1712-07107}.
This vulnerability can cause a series of problems, one of which is, whether we can manipulate a dialogue agent to say what we want.
In other words, can we find a well-designed input, to induce the agent to provide a desired output?
If this is possible, people with ulterior motives may take advantage of this weakness of the chatbots to guide them say something malicious or sensitive, causing adverse social impacts \cite{DBLP:journals/sigcas/WolfMG17,price2016microsoft,DBLP:journals/corr/abs-1809-04113}.

% \blue{I think for now this paragraph seems good}

% In order to answer this question, in this work, we imagine ourselves as these troublemakers, trying to design an algorithm that automatically crafts inputs such that lead a state-of-the-art black-box neural dialogue model to reply given targeted outputs. 
% This task is accompanied by two challenges.
In this paper, we want to study this dark side by seeking %answers 
an answer to the question -- whether we can design an algorithm that can automatically generate inputs that  %can 
lead a state-of-the-art black-box neural dialogue model to ``say what I want''.
However, this presents tremendous challenges.
% It presents tremendous challenges. \blue{the last sentence can be modified, e.g., However, this presents tremendous challenges.}
First, unlike similar works such as \cite{DBLP:conf/acl/HsiehYCZC18}, where the authors try to craft inputs for a neural image captioning model to output targeted sentences, our problem involves discrete inputs (i.e. texts rather than images) and treats the model as a black-box (since the setting is more realistic).
Thus, the traditional optimization method that finds the inputs by the guidance of gradient information is completely invalid.
Second, when trying to manipulate a dialogue system released by others, it is impractical for us to interact with it for unlimited times.
Based on this point, brute-force search methods cannot be adopted, and the number of the interactions with the black-box model that we need to find an input for a targeted output should be restricted to a reasonable level.
% \blue{I think maybe here at the start of this paragraph it can be mentioned that we focus our attention on the chatbots, or perhaps above, but it seems not a critical piece of information to mention so early in the paper.} 

% \blue{Also, rather than saying ``In order to answer this question, in this work, we imagine ourselves as these troublemakers, trying to design an algorithm that automatically crafts inputs such that lead a state-of-the-art black-box neural dialogue model to reply given targeted outputs. This task is accompanied by two challenges.'', we can perhaps say something like ``Now, from the perspective of the troublemakers and in view of the dark side, we seek to answer the question as to whether we can design an algorithm that automatically crafts inputs that can lead a state-of-the-art black-box neural dialogue model to ``say what I want''. In other words, we wish to act as a malicious puppeteer to guide the dialogue model output towards our targeted sentence, but naturally, this comes with tremendous challenges.'' In this way we are creating a more exciting story for the paper/work hopefully encouraging their interest to see the details.}

To address the above challenges, for a given black-box neural dialogue model, we propose to train a corresponding Reverse Dialogue Generator, which takes a targeted response as input and automatically outputs a query that leads the dialogue model to return that response.
The proposed Reverse Dialogue Generator is based on a Seq2Seq model and performs as a reinforcement learning (RL) agent.
The black-box dialogue model is regarded as the environment the agent interacts with.
% It regards the black-box dialogue model as the environment and interacts with it. \blue{this last sentence could be adjusted to fit better here, e.g., The black-box dialogue model is regarded as the environment the agent interacts with.}
It is optimized through policy gradients, with the similarity between the targeted outputs and what the dialogue model outputs with regard to a crafted input as the reward signal.
Extensive experiments conducted on a public well-trained neural dialogue model demonstrate the capacity of our model.

%We summarize our contributions as follows:

%\begin{itemize}
%\item We show the possibility of black-box neural dialogue models to be manipulated to ``say want I want'', which gives rise to the vigilance of the researchers towards this weakness of such models, and encourages further researches on how to defend them.
% \blue{maybe change to ``say what I want'' in quotes to relate to the current title }

%\item We propose an RL-based Reverse Dialogue Generator to effectively craft inputs that lead black-box neural dialogue models to output targeted responses.

%\item Experiments on a public well-trained black-box neural dialogue model show that the Reverse Dialogue Generator is able to find crafted inputs that lead to targeted responses with a considerable success rate, which demonstrate the effectiveness of the proposed model.
%\end{itemize}

\section{Related Works}

% {\bf we should not describe the work as ``[19] first investigates", typically, we will say "authors in [19]...." or ``xxxx is investigated in [19]", please change them in the related work first and then I will read this part after...}

% \red{Sure.}

Basically, our work is related to the problem of model attacks.
Although deep learning models have been used for many tasks that have shown to be useful across a plethora of domains, more recently, researchers have become aware to the fact that although these systems perform extremely well when in a perfect and stable environment (the type they were designed in), but when placed in the real world, they are quite easily susceptible to being attacked. Szegedy et al. \cite{DBLP:journals/corr/SzegedyZSBEGF13} first investigates the vulnerability of DNN-based image classifiers by crafting adversarial examples with imperceptible perturbations that lead the classifier to make mistakes.
Besides, Sharif et al. \cite{DBLP:conf/ccs/SharifBBR16} focus on attacking face recognition models; Xie et al. \cite{DBLP:conf/iccv/XieWZZXY17} try to find the weakness of an object detection system; in \cite{DBLP:conf/sp/KosFS18, DBLP:journals/corr/TabacofTV16}, researchers study the robustness of generative models, and novel methods to make adversarial examples for deep reinforcement learning based models are introduced in \cite{DBLP:conf/iclr/HuangPGDA17, DBLP:conf/iclr/KosS17}.

Adversarial attacks on deep learning models for NLP tasks also attract a lot of interest.
Attacking NLP models are more challenging since the inputs are discrete texts instead of continuous values such as image inputs \cite{DBLP:journals/corr/abs-1901-06796}. 
Text classification problems are studied in \cite{DBLP:journals/corr/abs-1812-00151, DBLP:conf/ijcai/0002LSBLS18}; sentiment analysis is involved in \cite{DBLP:conf/naacl/IyyerWGZ18}; grammar error detection is investigated in \cite{DBLP:conf/ijcai/SatoSS018}.
And Belinkov et al. \cite{DBLP:conf/iclr/BelinkovB18} try to fool a machine translation system, while Chan et al. \cite{DBLP:journals/corr/abs-1809-02444} study how to attack a neural reading comprehension model.
Besides, more works can be found in the survey \cite{DBLP:journals/corr/abs-1901-06796}.

For dialogue generation task, Wieting et al. \cite{DBLP:journals/corr/WietingBGL15a} explore the over-sensitivity and over-stability of neural dialogue models by using some heuristic techniques to modify original inputs and observe the corresponding outputs.
% Their work focused on the effects that take place when retraining the dialogue model using these adversarial examples to improve the robustness and performance of the underlying model, instead, in our work we are seeking to understand how well existing pre-trained dialogue models can be manipulated.
They evaluate the robustness of dialogue models by checking whether the outputs change significantly after the modifications on the inputs but don't consider targeted outputs.
The work which is most related to ours is \cite{DBLP:journals/corr/abs-1809-04113}, where the authors try to find trigger inputs which can lead a neural dialogue model to output a list of targeted egregious responses.
Different from our work, they treat the dialogue model as a white-box and take advantage of the model structure and parameters.
By comparison, our black-box setting is more challenging.
Furthermore, their algorithm indeed fails to lead the dialogue model to output the exact targeted responses.
It should be pointed out that similar with \cite{DBLP:journals/corr/abs-1809-04113}, our work is not a model attack task in the true sense since we only focus on the requirements of the outputs but don't force the inputs to be close to any original inputs.
However, our problems should be solved following the same ideas as adversarial attack problems.

By the way, nevertheless Cheng et al. \cite{DBLP:journals/corr/abs-1803-01128} don't focus on dialogue generation, they also try to manipulate a seq2seq model to generate texts with certain restrictions.
However, their non-overlapping and targeted keywords settings are looser than ours where a whole targeted response is required to be output, and this work is under the white-box assumption.

\section{Reverse Dialogue Generator}
\begin{figure}[t]
\begin{center}
\includegraphics[scale=0.4]{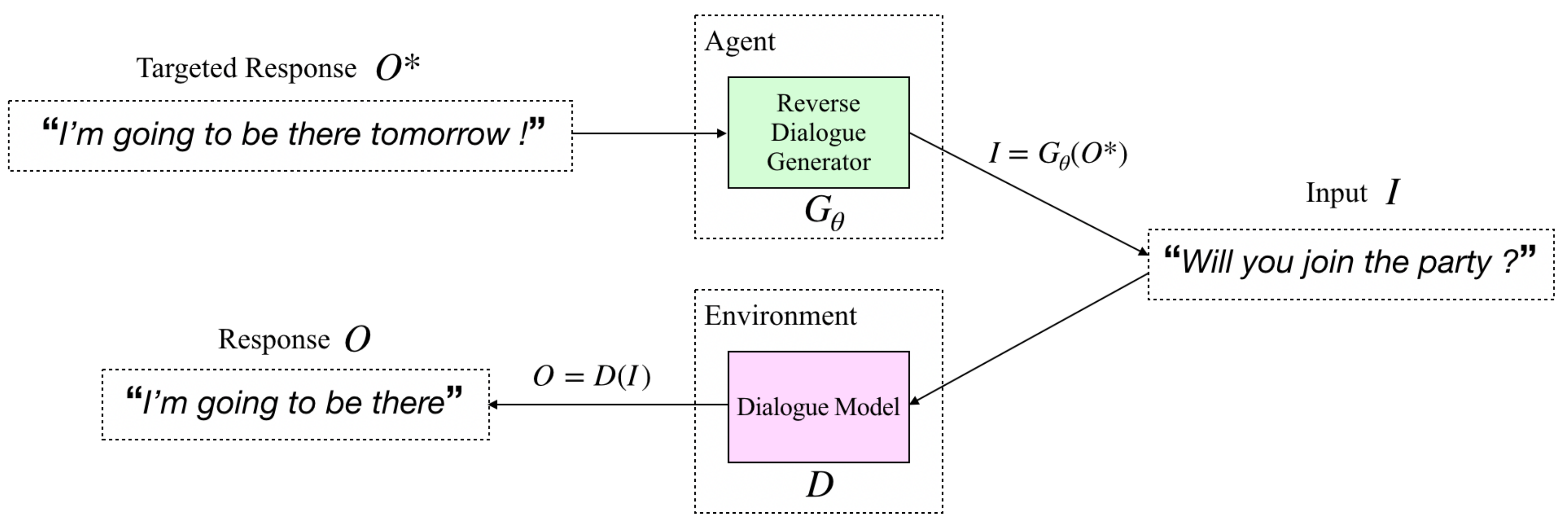}
\end{center}
%\vspace{-2ex}
\caption{The agent-environment setup of the proposed framework.} 
\label{fig:setup}
%\vskip -3ex
\end{figure}

In this work, we consider the specific neural dialogue model $D$ of our interest as a black-box environment.
The dialogue model $D$ is able to take an input sentence $I$ and output a corresponding response sentence $O$.
Now given a targeted output $O^*$, our goal is to find a well-designed input $I$, which leads the dialogue model $D$ to output a response $O$ that is exactly the same as $O^*$ ($O=O^*$), or at least similar with $O^*$ in semantics ($O\approx O^*$).
To achieve it, we build a Reverse Dialogue Generator agent $G_\theta$, which takes the targeted output $O^*$ as input to predict its corresponding input $I$.
A sketch of this agent-environment setup is shown in Figure \ref{fig:setup}. 
% \blue{this figure id is mentioned multiple times? to get Figure 5.1.2?}

\subsection{The Dialogue Model Environment}
In this work, we adopt the classical Seq2Seq neural dialogue model \cite{DBLP:journals/corr/VinyalsL15} as the dialogue model environment.
In the model, the encoder and the decoder are implemented by two independent RNN structures or its variants such as LSTM \cite{hochreiter1997long} and GRU \cite{DBLP:conf/ssst/ChoMBB14}.
% \blue{citations for these?}
The encoder reads the input sentence word by word and encodes it as a low-dimensional latent representation, which is then fed into the decoder to predict the corresponding response sentence word by word.
We assume the environment is a black-box; once the model is well-trained, we keep it sealed and the agent has no access to its structures, parameters or gradients.

Thanks to its strong power in the simulation and generalization of human conversations, Seq2Seq-based neural dialogue generator has been applied as an important component of practical chatbot services \cite{DBLP:journals/corr/abs-1812-08989}.
Thus, it is representative to be regarded as the object of study.
However, due to the black-box setting of the dialogue model environment, we can manipulate other neural dialogue models such as HRED \cite{DBLP:conf/aaai/SerbanSBCP16}, VHRED \cite{DBLP:conf/aaai/SerbanSLCPCB17}, etc., indiscriminately through our proposed method. We leave the investigation into other choices as one future work.

\subsection{The Reverse Dialogue Generator Model}
In order to find a corresponding input for a given targeted output, it is intuitive to formulate this problem as an optimization problem where we need to find an optimal input $I$ so that the similarity between the output $O=D(I)$ and the targeted output $O^*$ are maximized.
However, the standard gradient-based back-propagation approach is not applicable to this scenario.
Because the input $I$ consists of discrete tokens instead of continuous values, we cannot directly use gradient information to adjust it.
If we make adjustments with regard to the word embeddings, this method will generate results that cannot be matched with any valid words in the word embedding space \cite{DBLP:journals/corr/abs-1901-06796}.
What's more, in our black-box setting, we are not allowed to get the gradient information from $D$.

Therefore, in our work, we formulate this problem as a reinforcement learning problem.
A generative agent $G_\theta$, which is called Reverse Dialogue Generator, is trained to craft an input $I$ for a given targeted output $O^*$, with the aim to maximize the reward, that is, the similarity between the current output $O$ and the targeted output $O^*$.
The agent $G_\theta$ treats the input generation process as a decision-making process, and it is trained through policy gradient with the guidance of the reward signals.

An RNN language model can be adopted as the agent $G_\theta$.
However, we have to retrain it for each targeted output to obtain the desired input, which requires a great number of interactions with the $D$ for every single targeted output.
It is unachievable in reality when hackers try to manipulate a chatbot service.
Instead, we adopt a classical Seq2Seq structure as the agent $G_\theta$, which takes a targeted output as input and crafts the desired input.
It is trained offline, and when deployed in practical applications, it is able to generate the corresponding inputs for lots of targeted outputs automatically.
The details of its training process are described in Section \ref{sec:4.2}.

\section{Training}
In this section, we detail the training process of the dialogue model environment $D$ and the Reverse Dialogue Generator model $G_\theta$.

\subsection{The Dialogue Model Environment}
The Seq2Seq dialogue model is trained on a large-scale human conversation data collected on Twitter in a supervised manner with the aim to minimize the negative log-likelihood of the ground truth sentence given
inputs.
Afterward, it is treated as the black-box environment and its parameters are not further updated.
The details of the implementation can be found in Section \ref{sec:5.1.1}.

\subsection{RL Training of Reverse Dialogue Generator Model}
\label{sec:4.2}
In this subsection, we detail the RL training process of the proposed Reverse Dialogue Generator agent $G_\theta$.
In summary, the agent (Reverse Dialogue Generator $G_\theta$)  interacts with the environment (the dialogue model $D$).
Given a state (the targeted output $O^*$), the agent takes an action (generating an input $I$) following a policy $\pi_\theta$ (defined by the Seq2Seq model in the Reverse Dialogue Generator), and then receives a reward (the similarity between the targeted output $O^*$ and current output $O$) from the environment.
Afterward, the policy $\pi_\theta$ is updated to maximize the reward.

Next, we will introduce the environment, state, policy, action, and reward in detail.

\subsubsection{Environment}
The environment is the black-box dialogue model $D$.
When fed into an input $I$, it returns an output $O=D(I)$.
The input and the output are two dialogue utterances, which consist of a sequence of words with variable length.
\subsubsection{State}
A state is denoted as the targeted output $O^*$, that is, the input of the Reverse Dialogue Generator.

\subsubsection{Policy and Action}
The policy $\pi_\theta$ is defined by the Seq2Seq model in the agent $G_\theta$ and its parameters.
The Seq2Seq model forms a stochastic policy which assigns a probability to any possible input $I=(\omega_1, \dots, \omega_{T})$:

\begin{equation}
    \pi_\theta(I|O^*) = \prod_{t=1}^{T} P(\omega_t|\omega_1, \dots, \omega_{t-1}, O^*)
    \label{equ:1}
\end{equation}
where $T$ is the length of the input and $\omega_t$ is the $t$-th word.

The action is defined as the input $I$ to generate.
When observing the state $O^*$, the agent generates an input based on the distribution predicted by $\pi_\theta$ in Equation (\ref{equ:1}).
In the training phase, the input is chosen by stochastic sampling. And in the test phase, the input can be chosen in a greedy manner or through beam search.

\subsubsection{Reward}
Recall that our goal is to train the agent $G_\theta$ to craft an input for the dialogue model $D$ to return an output that is as similar with the targeted output $O^*$ as possible.
Let $O$ be the actual output returned by the dialogue model $D$ given a crafted input $I$.
We directly use the similarity between the targeted output $O^*$ and the current output $O$ as the reward for the input $I$ selected by $\pi_\theta$.

% Multiple metrics from word-overlap metrics such as BLEU, METEOR and ROUGE, to embedding based metrics, can be adopted as the measurement of the similarity.
% Compared with embedding based metrics, word-overlap metrics are more unilateral since they only consider the co-occurrence of the words in two sentences rather than contrasting them in the semantics level.
% Thus, we adopt the embedding average metric to measure the similarity.
We adopt the embedding average metric to measure the similarity.
This metric has been frequently used in many NLP domains, such as textual similarity tasks \cite{DBLP:journals/corr/WietingBGL15a}, since it's able to measure the similarity of two sentences in semantic level rather than simply consider the amount of word-overlap.
The embedding average approach first computes the sentence-level embedding $\bar e_r$ of a sentence $r$ by taking the average of the embeddings $e_\omega$ of all the constituent words $\omega$ in it: $\bar e_r = \frac{\sum_{\omega \in r} e_\omega}{\|\sum_{\omega' \in r} e_{\omega'}\|_{2}}$, and then the similarity between two sentences is defined as the cosine similarity of their corresponding sentence-level embeddings.
% \begin{equation}
%     \bar e_r = \frac{\sum_{\omega \in r} e_\omega}{\|\sum_{\omega' \in r} e_{\omega'}\|_{2}}
% \end{equation}
Given a crafted input $I$ and the targeted output $O^*$, we formally define the reward as the similarity between the current output $O$ and the targeted output $O^*$:
\begin{equation}
    \label{eq:l_reward}
     R(I|O^*) = {\rm Sim}(O^*, D(I)) = {\rm cos}(\bar e_O, \bar e_{O^*})
\end{equation}

\subsubsection{Optimization}
With the reward function defined above, the objective function that the agent aims to maximize can be formulated as follows:
\begin{equation}
    \label{eq:obj_1}
     J(\theta) = \mathbb{E}_{I\sim \pi_\theta(I|O^*)} R(I|O^*) 
\end{equation}
The accurate value of $J(\theta)$ in Equation (\ref{eq:obj_1}) is very difficult to obtain in practice. Therefore, previous works have proposed many methods to estimate it and its gradient, which is then used to update the parameters $\theta$ of the policy (i.e., $\pi_{\theta}$).

To optimize the objective in Equation (\ref{eq:obj_1}), we apply the widely used REINFORCE algorithm \cite{williams1992simple}, where Monte-Carlo sampling is applied to estimate $\nabla_\theta J(\theta)$. Specifically, 
\begin{align}
    \label{eq:gradient_1}
    \nabla_\theta J(\theta) &= \sum_{I} R(I|O^*) \nabla \pi_\theta(I|O^*) \\ \nonumber
    & = \sum_{I} R(I|O^*) \pi_\theta(I|O^*) \nabla \log\pi_\theta (I|O^*)\\\nonumber
    & = \mathbb{E}_{I\sim\pi_\theta(I|O^*)}[R(I|O^*) \pi_\theta(I|O^*) \nabla \log\pi_\theta (I|O^*)] \\\nonumber
    & \approx \frac{1}{N}\sum^N R(I|O^*) \pi_\theta(I|O^*) \nabla \log\pi_\theta (I|O^*) \nonumber
\end{align}

With the obtained gradient $\nabla_\theta J(\theta)$, the parameters of the policy network $\pi_\theta$ can be updated as follows:
\begin{align}
    \label{eq:update}
    \theta := \theta + \alpha  \nabla_\theta J(\theta)
\end{align}
where $\alpha$ is the learning rate.
Thus, the REINFORCE algorithm for updating the policy $\pi_\theta$ can be summarized as: For each targeted output $O^*$, we first sample $N$ inputs according to the distribution $\pi_\theta(\cdot|O^*)$.
Then we estimate the rewards of the sampled inputs and calculate the gradient.
Finally, we update the parameters of the policy network.

\section{Experiments}
We conduct extensive experiments to evaluate the effectiveness of the proposed Reverse Dialogue Generator.
We measure the success rates of the proposed model in seeking out the desired inputs towards various targeted outputs and explore the performance of the model under various settings.
In this section, we will go into details about the experimental settings and results.

\subsection{Experimental Settings}
% In this subsection, we present the details of the experimental settings.
% We will first introduce the dialogue model under study and then elaborate on the specific implementation of our proposed model.

\subsubsection{The Dialogue Model}
\label{sec:5.1.1}
To ensure the reproducibility of the experiments, we directly adopt the well-trained Seq2Seq dialogue model released on the dialogue system research software platform ParlAI \cite{DBLP:conf/emnlp/MillerFBBFLPW17} as the dialogue model environment.
The implementation of the dialogue model is detailed as follows.
In the Seq2Seq model, both the encoder and the decoder are implemented by 3-layer LSTM networks with hidden states of size 1024.
As the standard practice, the initial hidden state of the decoder is set as the same as the last hidden state of the encoder.
The vocabulary size is 30,000.
Pre-trained Glove word vectors \cite{pennington2014glove} are used to initialize the word embeddings whose dimension is set as 300.
The model had been trained through stochastic gradient descent (SGD) with a learning rate of 1.0 on 2.5 million Twitter single-turn dialogues. In the training process, the dropout rate and gradient clipping value are both set to be 0.1.
It should be pointed out again that in the following experiments, this dialogue model is treated as a black-box which takes an input sentence and output a response sentence.

\subsubsection{The Reverse Dialogue Generator}
\label{sec:5.1.3}
As described above, we adopt a Seq2Seq structure as the Reverse Dialogue Generator.
Two 2-layer LSTM networks with the hidden size of 1,024 are applied as the encoder and the decoder respectively.
The vocabulary size is set to be 60,000, and all the size of word embeddings is 300.
The word embeddings are randomly initialized and fine-tuned during the training process.
The last hidden state of the encoder is treated as the context vector which is used to initialize the hidden state of the decoder.
% The model is implemented with Pytorch~\footnote{https://pytorch.org/}.

\textbf{Pre-training.}
Before RL training, we first initialize the agent by pre-training it on output-input pairs in a reference corpus in a supervised learning manner.
We build the reference corpus in this way: we first randomly collect 160K posts from Twitter, and then we feed them into the dialogue model to get the corresponding responses output by the model.
Finally we obtain 160K output-input pairs as the reference corpus.
Specifically, in the pre-training process, the model is optimized by the standard SGD algorithm with the initial learning rate of 20.
At the end of each epoch, if the loss doesn't decrease in the validation set, the learning rate is reduced with a decay rate of 0.25.
And the batch size is 16.
In addition, to prevent overfitting issues, we apply the dropout with the rate of 0.1 and gradient clipping with clip-value being 0.25.

\textbf{RL training.}
In the RL training process, all the parameters in the model are optimized as Equation (\ref{eq:update}) through Adam optimizer with an initial learning rate of 0.001.
The 160K outputs in the reference corpus are used as the targeted outputs for RL training.
The batch size, dropout rate, and gradient clipping value are set the same as those in the pre-training process.
When calculating the rewards (i.e. the embedding similarities), we only consider the tokens out of a fixed stopword list which consists of common punctuations and special symbols.
Pre-trained Google news word embeddings\footnote{https://code.google.com/archive/p/word2vec/} are adopted to compute the similarities.
And in order to accelerate the convergence of the model, we set all the rewards less than 0.5 to be 0, and the others remain the original values.

\textbf{Decoding.}
In the test phase, the greedy method and beam search can be used for decoding.
And we empirically find that if we first get $N$ candidates with top-$N$ scores in beam search and then feed them into the dialogue model, treat the one with the highest reward as the crafted input, the performance improves significantly.
In the experiments, we report the results of greedy decoding of the pre-trained model (Pre-trained Greedy), greedy decoding of the RL-trained model (RL Greedy), and beam search of the RL-trained model with $N$ candidates (RL BeamSearch($N$)).

\begin{figure}[t]
\begin{center}
\includegraphics[scale=0.4]{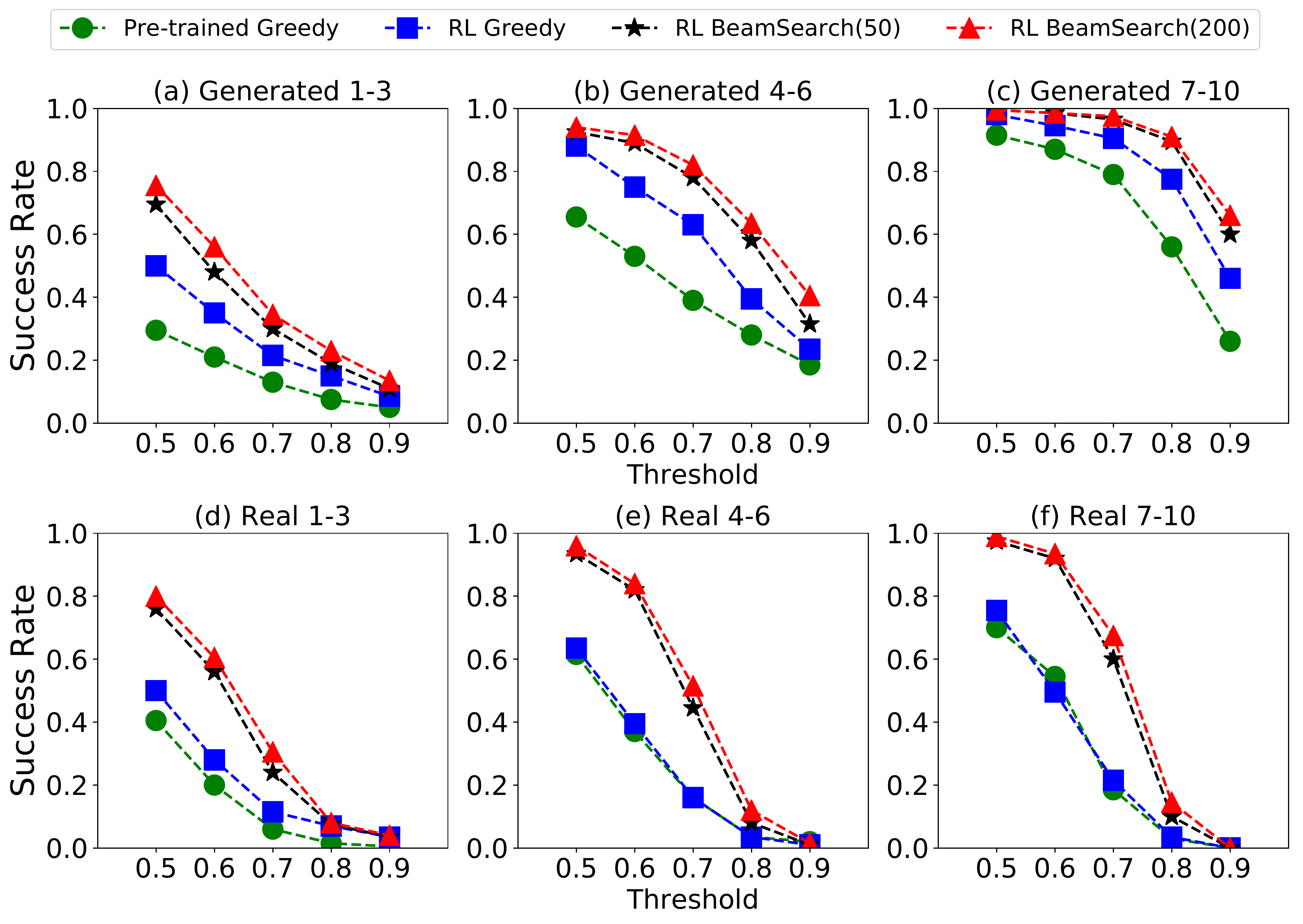}
\end{center}
%\vspace{-2ex}
\caption{Success rates of the pre-trained model and the RL-trained model with different decoding methods. The upper row and the lower row show the results on the Generated target list and the Real target list with various lengths respectively.}
\label{fig:res}
%\vskip -3ex
\end{figure}

\subsection{Experimental Results}
\label{sec:5.2}
We conduct experiments on two types of target output lists.
To construct the \textbf{Generated} target output list, we first feed 10k human utterances (have no overlap with the reference corpus) from Twitter into the dialogue model to get 10k generated responses and then randomly sample 200 responses as the targets in length 1-3, 4-6 and 7-10, respectively.
The \textbf{Real} target output list is obtained by randomly collecting sentences directly from Twitter.
The number of target outputs in each length group is also 200.

Given a targeted output, when the proposed model finds an input which leads to an output whose similarity to the targeted one is above a preset threshold, we say the manipulation is successful.
Figure~\ref{fig:res} shows the success rates of our proposed model for manipulating the Twitter dialogue model in two experimental settings.
The figures show how success rates vary with different thresholds.
% {\bf we do not define when we are successful?? please explicitly define it in the experimental setting section???}
First of all, from the figures, we can see that for both the \textbf{Generated} and the \textbf{Real} target lists, RL-based model with beam search can achieve a success rate around $80\%-100\%$ with a score $0.5$.
Especially for more than around $80\%$ \textbf{Generated} targets with length greater than or equal to $4$, we can find desired inputs that lead to a similarity score above $0.7$. 
Second, we can see that in most of the scenarios, RL Greedy obviously outperforms Pre-trained Greedy, which demonstrates that using reward signals to fine-tune the policy network through RL significantly strengthens its capability for crafting desired inputs.
Besides, compared with greedy decoding method, beam search with multiple candidates improves the success rates a lot.
And compared with $N=50$, the beam search methods with $N=200$ candidates improve the performance slightly, which means that interacting with the dialogue model for a reasonably small number of times can guarantee a considerable success rate.
Another key observation is that the model performs significantly better on the \textbf{Generated} target list than on the \textbf{Real} target output list.
Actually, the neural dialogue models suffer from the safe response problem \cite{DBLP:conf/naacl/LiGBGD16}. Such models tend to offer generic responses to diverse inputs, which makes it hard for the model to provide a specific targeted response (often seen in real human conversations).

\begin{table}
  \caption{Average embedding similarity scores between the output and the target output in terms of \textbf{Real} target output list.}
  \label{tab:ave_scores}
  \centering
  \begin{tabular}{c|ccc}
    \toprule
    \cmidrule(r){1-2}
    \textbf{Length}     & 1-3     & 4-6 & 7-10 \\
    \midrule
    Real Input & 0.439  & 0.518 & 0.566     \\
    \hline
    Pre-trained Greedy & 0.446  & 0.529 & 0.559     \\
    \hline
    RL Greedy & 0.486  & 0.560 & 0.588     \\
    \hline
    RL BeamSearch(50) & 0.599  & 0.678 & 0.709     \\
    \hline
    RL BeamSearch(200) & 0.621  & 0.694 & 0.726     \\
    \bottomrule
  \end{tabular}
\end{table}

In order to further demonstrate the effectiveness of the proposed framework, for each \textbf{Real} targeted output, we feed its corresponding real inputs in the corpus into the dialogue model to check how similar the output responses and the target ones are.
We calculate these similarity scores for each \textbf{Real} targeted output for the real inputs and the inputs found by the proposed model and report the average value in Table~\ref{tab:ave_scores}. 
According to the table, we observe that even inputting the real inputs, the similarity scores between the outputs and the target outputs are not high.
Besides, with the crafted inputs from the proposed framework, these similarity scores are significantly improved.
For example, for RL BeamSearch(200), the similarity is improved by $41.5\%$, $34.0\%$ and $28.3\%$ for the target outputs with length 1-3, 4-6 and 7-10, respectively.

\begin{table*}[!t]
\centering
  \caption{Case Study. The first column shows the inputs found by RL BeamSearch(200) according to given target outputs. The middle column shows the target outputs and the outputs generated by the dialogue model to the inputs. The column on the right side shows the embedding similarity score between outputs and target outputs.}
  \begin{tabular}{l|l|l}
  \hline
    \textbf{Inputs} & \textbf{Responses} & \textbf{Similarity}\\
     \hline
     \multirow{3}{4.5cm}{soo calm . you should be nervous .} &  \textbf{Target}: i ' m just trying to be a good person &\\
     & \textbf{Output}:  i ' m not . i ' m just trying to be a & 0.952\\
     &  better person . &\\
     \hline
     \multirow{2}{4.5cm}{guess she'll be invited .} &  \textbf{Target}: i ' m sure she ' ll be fine . & {0.946}\\
     & \textbf{Output}: i ' m sure she ' ll be a good one . & \\
     \hline
     \multirow{2}{4.5cm}{neither ready pls} &  \textbf{Target}: i ' m not ready for this & 1.0\\
     & \textbf{Output}:  i ' m not ready for this & \\
     \hline
     \multirow{2}{4.5cm}{how is nephew ?} &  \textbf{Target}: he ' s a good guy & 0.982\\
     & \textbf{Output}: he ' s good . he ' s a good guy . & \\
     \hline
     \multirow{2}{4.5cm}{you weren't invited .} &  \textbf{Target}: i was there & 1.0\\
     & \textbf{Output}: i was there . & \\
     \hline
    
  \hline
  \end{tabular}
  \label{tab:case}
    \vskip -2ex
\end{table*}

Table~\ref{tab:case} shows five examples in the manipulating experiments.
The first three target outputs are from the \textbf{Generated} target output list, while the other two are from the \textbf{Real} target list.
Given those target outputs, desired inputs are successfully crafted.
Each of them leads to an output of the dialogue model similar or equal to the target one, evaluating by the embedding similarity measurement.
Besides, unlike some related works \cite{DBLP:journals/corr/abs-1809-04113, DBLP:journals/corr/abs-1803-01128} where crafted text inputs are ungrammatical and meaningless, the inputs generated by our model are smooth and natural utterances.
This is because the decoder of the Seq2Seq model in the Reverse Dialogue Generator serves as a language model, which guarantees the smoothness of the generated inputs.

% \subsubsection{The Twitter Corpus}
% A public Twitter corpus\footnote{"twitter\_en big" in the repository https://github.com/Marsan-Ma/chat\_corpus/} which contains around 5M tweets scraped from Twitter is used in our experiments to build the reference corpus (see Section \ref{sec:5.1.3}), target output lists (see Section \ref{sec:5.2}), etc.
% We note that this corpus is different from the dialogue dataset that is used to train the dialogue model.

% In order to train the Reverse Dialogue Generator and build the target output lists (see Section \ref{sec:5.2}), we use real human dialogue data from a public Twitter corpus which contains around 5M tweets scraped from Twitter.
% We note that this corpus is different from the dialogue dataset that is used to train the dialogue model.

% \footnote{"twitter\_en big" in the repository https://github.com/Marsan-Ma/chat\_corpus/}

% \footnote{https://parl.ai/docs/zoo.html\#twitter-conversational-model}

\section{Conclusion}

Recently, dialogue systems are being integrated into our daily lives at a quite rapid pace.
In the practical implementation of dialogue systems, neural dialogue models play an important role.
However, recent concerns have risen for neural models across all domains as to whether they can be manipulated (in most cases, by crafted adversarial examples), which inspires us to examine the same problem of neural dialogue models.
Our work reveals a dark side of such models that they are likely to be manipulated to "say what I want" -- by finding well-designed inputs, we can induce the dialogue agent to provide desired outputs.
We propose a reinforcement learning based \textbf{Reverse Dialogue Generator} which learns to craft such inputs automatically in the process of interacting with the black-box neural dialogue model.
Extensive experiments on a representative neural dialogue model demonstrate the effectiveness of our proposed model and show that dialogue systems used in our daily lives can indeed be manipulated, which is a warning about the security of dialogue systems for both the research community and the industry.

% In addition, our proposed method is not only able to manipulate neural dialogue model, but it's also likely to be applied on black-box dialogue systems based on other methods (e.g. rule-based, retrieval-based, etc.), or even models for other natural language generation tasks (e.g. text summarization, machine translation, etc.).
% We will leave the investigations on these areas as future works.
For future works, we plan to extend the current framework to other sequence models.
Besides, in this work, we examine the security problem of dialogue systems.
In future works, we will also investigate concerns about the privacy of them, specifically, the possibility of dialogue systems to leak the sensitive information of users.

\bibliographystyle{unsrt}
\bibliography{main}

\end{document}